%% file: coling_latex.tex
\pdfoutput=1

\documentclass[11pt]{article}

\usepackage[preprint]{coling}

\usepackage{times}
\usepackage{latexsym}

\usepackage[T1]{fontenc}

\usepackage[utf8]{inputenc}

\usepackage{microtype}

\usepackage{inconsolata}

\usepackage{graphicx}
\usepackage{amsfonts}
\usepackage{amsmath}
\usepackage{bm}
\usepackage{float}
\usepackage{amsthm}
\theoremstyle{definition}

\usepackage{algorithm}
\usepackage{algpseudocode}
\usepackage{listings}
\usepackage{xcolor}
\usepackage{stfloats}
\definecolor{codegreen}{rgb}{0,0.6,0}
\definecolor{codegray}{rgb}{0.5,0.5,0.5}
\definecolor{codepurple}{rgb}{0.58,0,0.82}
\definecolor{backcolour}{rgb}{0.98,0.98,0.95}
\lstdefinestyle{mystyle}{
    commentstyle=\color{codegreen},
    keywordstyle=\color{magenta},
    numberstyle=\tiny\color{codegray},
    stringstyle=\color{codepurple},
    basicstyle=\ttfamily\footnotesize,
    breakatwhitespace=false,
    breaklines=true,
    captionpos=b,
    keepspaces=true,
    numbersep=5pt,
    showspaces=false,
    showstringspaces=false,
    showtabs=false,
    tabsize=2
}
\usepackage{enumitem}
\newtheorem{assumption}{Assumption}
\DeclareFixedFont{\ttb}{T1}{txtt}{bx}{n}{12} 
\DeclareFixedFont{\ttm}{T1}{txtt}{m}{n}{12}  

\title{A Combinatorial Approach to Neural Emergent Communication}

\makeatletter
\def\@fnsymbol#1{\ensuremath{\ifcase#1\or \dagger\or \ddagger\or
   \mathsection\or \mathparagraph\or \|\or **\or \dagger\dagger
   \or \ddagger\ddagger \else\@ctrerr\fi}}
\makeatother

\author{Zheyuan Zhang\thanks{Work done at SLED Lab, University of Michigan} \\
  University of Michigan \\
  Ann Arbor, MI \\
  \texttt{zheyuan@umich.edu} \\}

\begin{document}
\maketitle
\begin{abstract}
Substantial research on deep learning-based emergent communication uses the referential game framework, specifically the Lewis signaling game, however we argue that successful communication in this game typically only need one or two symbols for target image classification because of a sampling pitfall in the training data. To address this issue, we provide a theoretical analysis and introduce a combinatorial algorithm SolveMinSym (\texttt{SMS}) to solve the symbolic complexity for classification, which is the minimum number of symbols in the message for successful communication. We use the \texttt{SMS} algorithm to create datasets with different symbolic complexity to empirically show that data with higher symbolic complexity increases the number of effective symbols in the emergent language.
\end{abstract}

\input{sections/intro}
\input{sections/related_works}
\input{sections/main}
\input{sections/experiments}
\input{sections/conclusions}

\section{Limitations}
The data distribution of ${\min}(|M|)$ is highly narrow so it is difficult to collect data with high ${\min}(|M|)$ (e.g. $>3$). Therefore, our experiments only compare ${\min}(|M|) = 2$ with ${\min}(|M|) = 3$. Future works should explore how to synthetically generate data with arbitrary ${\min}(|M|)$ directly.

\section{Acknowledgements}
The author would like to thank Ziqiao Ma, Joyce Chai for their helpful discussions and the anonymous reviewers for
their valuable comments and suggestions.

\bibliography{custom}

\appendix
\input{sections/appendix}

\end{document}

%% file: sections/intro.tex
\section{Introduction}

In a multi-agent environment, communication often naturally evolves as a strategic behavior \cite{russell2010artificial}. Emergent communication is in such setting typically, in which the communication channel between agents gets optimized for solving a cooperative task, e.g. Lewis signaling games \cite{lewis1969convention}. However, we argue that the languages emerged from these games are in a simple form of language (e.g. one or two words), differs from the human language, underlying complexity from compositionality. Therefore, in this paper, we investigate the theoretical effective symbols for the emerged language in Lewis signaling games.


%% file: sections/related_works.tex
\section{Related Work}
\label{sec:other_related_works}
The Partially Observable Markov Decision Process (POMDP) extends the Markov Decision Process (MDP) framework in which the agent is not able to access the complete state information of the environment \cite{lovejoy1991survey} and communication potentially leads to coordination of agents for higher rewards. Also, some scholars in linguistics, psychology and biology posit that language evolved from primate communication \cite{tomasello1997primate, tomasello2013cultural, pika2006referential}.

Many works in emergent communication and language emergence use the referential game framework which is the Lewis signaling game, providing a setting for learning communication protocol, as well as the analysis of the emergent language \cite{lewis1969convention, lazaridou2016multi, havrylov2017emergence, evtimova2018emergent, bouchacourt2018agents, bouchacourt-baroni-2019-miss, kharitonov2019egg, michel2022revisiting}. There are also works explored other games \cite{mordatch2018emergence, das2019tarmac, mu2021emergent}, intrinsic motivations and extrinsic environmental pressures for language emergence \cite{chentanez2004intrinsically, cornudella2015intrinsic, gaya2016role, hazra2020intrinsically, hazra2021zero, li2019ease, cogswell2019emergence}. More recent works explore the role of emergent language for embodied AI and robotics \cite{liu2023simple, mu2023ec2}. We refer to Appendix~\ref{appendix:ext_related_works} for more related works.

The work most related to this paper is \citet{kottur-etal-2017-natural}. Their game is a Task \& Talk reference game which has multiple rounds of dialog. A-BOT is given an object unseen by Q-BOT and Q-BOT is assigned a task consisting of two attributes. The goal is to find these two attributes of the hidden object for Q-BOT through dialog with A-BOT. They found that overcomplete vocabularies result in no dialog, instead having a codebook that maps symbols to objects. Our paper, differs from this work, instead of using a multi-round dialog game, we use the classical Lewis signaling game. Additionally, we mainly investigate the emergence of longer compositional language.

%% file: sections/main.tex
\section{Symbolic Complexity for Classification}
For this paper, we consider the classification-based Lewis signaling game or referential game introduced by \citet{lazaridou2016multi, havrylov2017emergence}:

\begin{enumerate}[leftmargin=*]
    \setlength\itemsep{-0.25em}
    \item A target image $I_{t}$, along with K distracting images $\{I_{d_k}\}^K_{k=1}$ are sampled at random from a set of images.
    \item Two agents, a sender $S_{\theta_1}$ and a receiver $R_{\theta_2}$.
    \item The sender sends a message $M$ to the receiver after observing the target image.
    \item The receiver's objective is to select the target image from $\{I_{t}, I_{d_1}, I_{d_2}, \ldots, I_{d_K}\}$ given $M$.
\end{enumerate}

The communication is successful if the receiver correctly selects the target image from distracting images. We denote $|M|$ as the length of the message $M$ and $L$ as the maximum message length for communication. The motivation of our work comes from observation in varying $|M|$ \cite{havrylov2017emergence}. We find that there's no significant improvement on communication success when increasing $L$ from 2 to 3 and subsequent increments. Therefore, $M$ requires only 2 tokens for successful communication. This means the number of effective symbols in $M$ is 2. In order to study how to emerge longer effective language, we use attribute-value vocabulary, similar to \citet{kottur-etal-2017-natural} and create synthetic datasets for theoretical analysis and more controlled experiments. Specifically, we define $|A|$ attributes ${a_1, a_2, a_3, ..., a_{|A|}}$, and $|V_{a_i}|$ values for each attribute $a_i$.

We argue that the bottleneck of producing longer effective $M$ lies on the \textbf{inherent limitations of the dataset} being used to train the agents. For example, suppose we sample 10 images, denote as $I_{\{1,2,3,...,10\}}$ and each $I_i$ represents a distinct object category (e.g. ball, bottle, person). In this case, we only need one symbol for classification of any target image, if each class is represented by 1 symbol. We have the following assumption.

\begin{assumption}
Any Lewis signaling game has a symbolic complexity (i.e. minimum number of symbols), ${\min}(|M|)$, for successful communication, i.e. correct classification of the target image.
\end{assumption}

\begin{figure}[!t]
    \centering
    \includegraphics[width=0.46\textwidth]{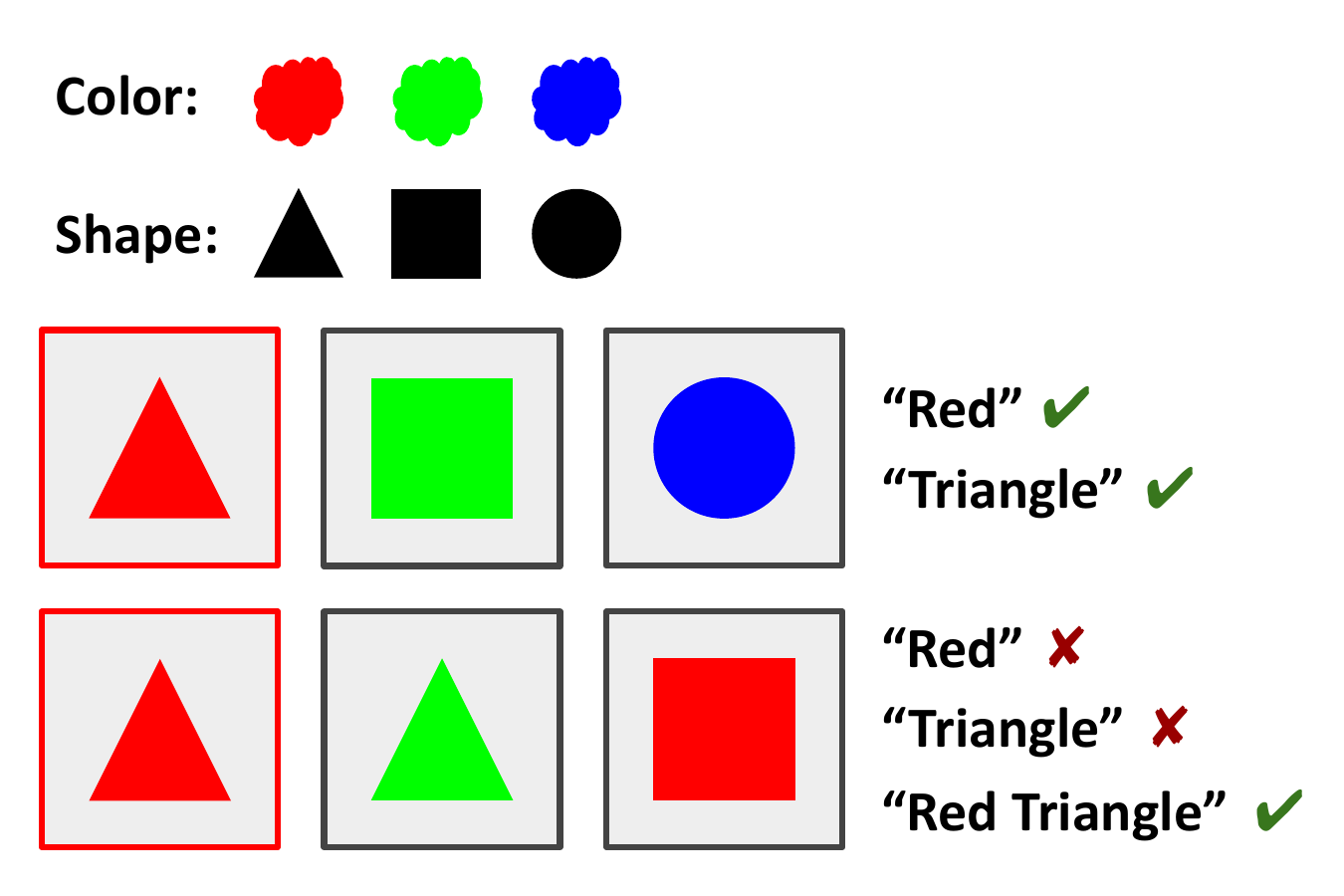}
    \vspace*{-10pt}
    \caption{Example of minimum symbols for successful classification. \vspace*{-10pt}}
    \label{fig:minsym_example}
\end{figure}

Figure \ref{fig:minsym_example} shows an example for identifying ${\min}(|M|)$ for successful classification of the target image in a synthetic setting. In this synthetic example, we consider 2 attributes, color and shape. Target images in both the top row and the bottom row are those with red boundaries and rest are distracting images. For the top row, we only need 1 symbol "Triangle" or "Red" to correctly discern the target images from distracting images. For the bottom row, 1 symbol is not enough because only "Triangle" matches to the second image which is a distracting image and only "Red" matches to the third image which is also a distracting image. Therefore we need at least 2 symbols "Red Triangle" for the successful classification. Therefore, ${\min}(|M|)=1$ for the top row and ${\min}(|M|)=2$ for the bottom row, despite that both rows have the same target image and same number of distracting images.

\subsection{The Pitfalls in the Lewis Signaling Game}
Using the example above, if ${\min}(|M|)$ > 1, it requires sampled images to have the same object category (e.g. ball) and the target image belongs to that object category. This setting forces the model to use another symbol to further distinguish the target image from distracting images (e.g. red ball, blue ball). If we have an attribute $a_1$ to represent "object category", and $|V_{a_1}|$ represents number of object categories, we randomly sample $n$ images where $n=K+1$ under the assumption that the dataset is uniformly distributed (i.e. $P(a_i)=P(a_j)$ for any $i$ and $j$) and we sample with replacement for simpler analysis, the probability of the sampled images to have at least $m$ images of the same class for any class $X$ is:
\vspace*{-5pt}
\begin{multline}
\label{eqn:1}
    P(X\geq m) = \\
    \sum^n_{k=m} \binom{n}{k} (\frac{1}{|V_{a_1}|})^k(1-\frac{1}{|V_{a_1}|})^{n-k}
\end{multline}
\vspace*{-3pt}
Since we are interested in any class achieving this count, the probability of at least one class has at least $m$ images of the same class is:
\vspace*{-3pt}
\begin{equation}
\label{eqn:2}
   P(\exists X\geq m) = 1-(1-P(X\geq m))^{|V_{a_1}|} 
\end{equation}
From Eqn.\ref{eqn:1} and Eqn.\ref{eqn:2}, as $n$ increases, $P(\exists X\geq m)$ increases. One could infer that if we increase the number of distracting images, it is likely needing more symbols for successful communication. However, this is rarely the case because as $|V_{a_1}|$ increases, $P(\exists X\geq m)$ decreases. The Microsoft COCO dataset, MSCOCO \cite{chen2015microsoft} used in \citet{havrylov2017emergence} contains 80 annotated object categories. In fact, since MSCOCO are real images, rather than synthetically generated images, the total number of object categories is (much) larger than 80. Therefore, even if $n$ becomes large, $P(\exists X\geq m)$ can still be small because $|V_{a_1}|$ is too large. \citet{havrylov2017emergence} uses 127 distracting images so $n=128$. However, for example, if $|V_{a_1}| = 10000$ and $m=2$ minimally, $P(\exists X\geq m) \approx 55\%$ before sampling the target image. In this case, for most of the training data, ${\min}(|M|) = 1$. This theoretical analysis is supported by the experimental findings in their work, which show that $|M| = 1$ results in a 60\% communication success rate. Additionally, even if images are in the same class, further distinguishing the target image mostly likely requires only 1 additional symbol.

\subsection{Combinatorial Algorithm for Solving Symbolic Complexity}
In the previous section, we analyzed that it would only require 1 to 2 symbols for successful classification in most real images (e.g. MSCOCO) for Lewis signaling game. Our hypothesis is that \textbf{increasing} \boldsymbol{${\min}(|M|)$} \textbf{for the data itself leading to emerge longer effective language}, as an alternative to designing new learning algorithms for agents. 

\paragraph{Dataset Generation}
In order to solve ${\min}(|M|)$, we synthetically generate a dataset for controlling attributes and values for each attributes, which is not available for real images. Previously, we defined $|A|$ attributes and $|V_{a_i}|$ values for each attribute $a_i$. Here, for simpler analysis and more controlled experiments, we consider a special case where $|V_{a_i}| = |V_{a_j}|$ for any $i$ and $j$. That is, number of values is the same for all attributes. Therefore, we can simplify the notation to $|A|$ attributes and $|V|$ values. 
Similar to \citet{li2019ease} and \citet{kottur-etal-2017-natural}, each image $I$'s representation is a vector $V_{I_i}$ concatenating one-hot vectors of attributes, that is, $V_{I_i} \in \mathbb{R}^{|A|\times |V|}$.
In this paper, we use $|A|=20$ and $|V|=4$.

\paragraph{\texttt{SolveMinSym (SMS)} Algorithm}
We now have access to the ground-truth attribute values of the image, then we use a combinatorial approach to solve the minimum number of symbols given the target image and all images. Generally, we generate all possible non-empty combinations with respect to the target image from least number of symbols (i.e. 1) to maximum number of symbols (i.e. $|A|$). Then, we iterate through these combinations for checking whether the iterated combination uniquely identifies the target image and return the length of the combination once it returns true. Unique identification determines whether the combination matches to any distracting image (i.e. making target image not unique which fails the communication). Theoretically, since the length of the combination increases from minimum to maximum, the ${\min}(|M|)$ equals to the length of the combination returned. We refer to Appendix~\ref{appendix:sms} for the implementation details.

\paragraph{\boldsymbol{${\min}(|M|)$} Controlled Sampling for Lewis Signaling Game}
To approach our hypothesis on increasing symbolic complexity for the training data itself leading to emerge longer effective language. We implemented the ${\min}(|M|)$ Controlled Sampling algorithm as below:

\setcounter{algorithm}{1}
\begin{algorithm}[H]
\caption{${\min}(|M|)$ Controlled Sampling}
\begin{algorithmic}[1]
\State \textbf{Input}: Dataset $D$, size of generated data $N_g$, number of ${\min}(|M|)$ $N_{min}$, number of distracting images $N_d$
\State $D_L$ = \{\}
\While{$|D_L[0...N_{min}]| < N_g$}
    \State $I \sim \text{Uniform}(D, N_d+1)$
    \State $I_t \sim \text{Uniform}(I, 1)$
    \State $N_{ms}$ = \texttt{SMS}($I_t$, $I$)
    \State Add ${I_t, I}$ to $D_L[N_{ms}]$
\EndWhile

\State \textbf{Return} $D_L$
\end{algorithmic}
\end{algorithm}

\vspace{-5pt}

By applying the controlled sampling algorithm to our synthetically generated data, we are able to generate data with different ${\min}(|M|)$. In this paper, we use $N_g = 10000$ (8000 used for training and 2000 used for evaluating), $N_d = 63$ (sample 64 images each time). By Eqn.\ref{eqn:1} and Eqn.\ref{eqn:2}, we find theoretically and also empirically that the ${\min}(|M|)$ data is a highly narrow distribution which the majority of data are on two different ${\min}(|M|)$s for a certain configuration of $|A|$ and $|S|$. Therefore, in order to collect 10000 datapoints within a reasonable time. We only have two sets of data with ${\min}(|M|) = 2,3$.

%% file: sections/experiments.tex
\section{Experiments}
We parameterize the sender $S_{\theta_1}$ and the receiver $R_{\theta_2}$ with GRU \cite{cho2014properties}. The hidden dimension of the GRU is 512 and the embedding size is 32. We use Schedule-Free AdamW for optimization \cite{defazio2024road}. The vocabulary size is set to be $|A| \times |V| = 20 \times 4 = 80$. We do not use an arbitrary large vocabulary size (e.g. 10000) because it is disadvantageous to compositionality as it encourages one symbol to represent composed phrase (e.g. Both "Red" and "Red Triangle" can be represented by only 1 symbol). We also do not use a minimal vocabulary size $|V|$ because it forces the model to learn the ordering of attributes and it is not realistic to human languages (e.g. we use "Red" and "Triangle" as two different words. It's hard to interpret a phrase like "Red Red" with the first word corresponds to the color and the second word corresponds to the shape).
The computation graph of $S_{\theta_1}$ contains sampling (i.e. generation) so it becomes nondifferentiable, therefore we use Gumbel-Softmax Relexation, as detailed in Appendix~\ref{appendix:gumbel}.

\begin{figure}[!t]
    \centering
    \includegraphics[width=0.43\textwidth]{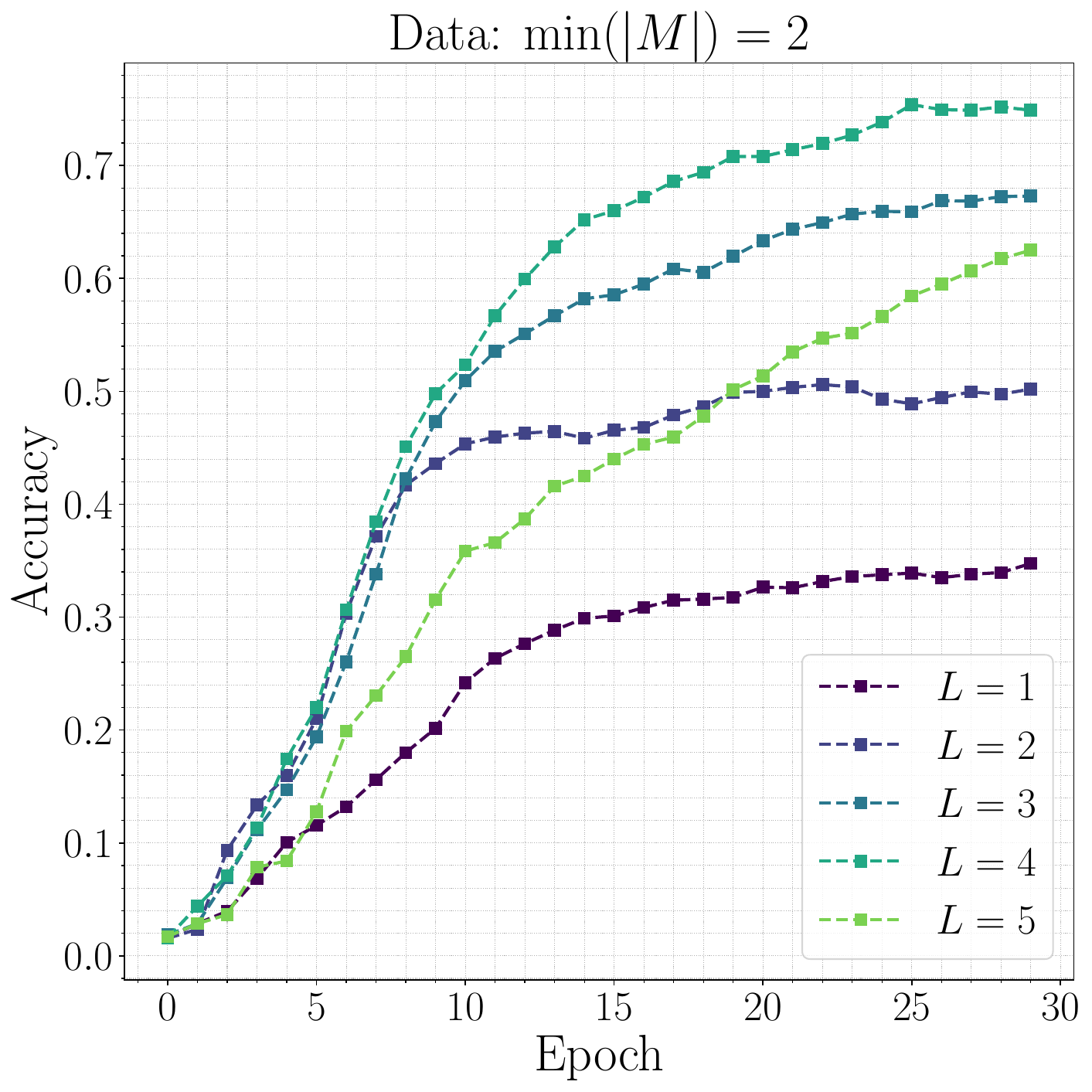}
    \vspace*{-10pt}
    \caption{Validation accuracy over epochs with different maximum message lengths ($L$) on data where $\min(|M|) = 2$. \vspace*{-10pt}}
    \label{fig:results_ms2}
\end{figure}
\begin{figure}[!t]
    \centering
    \includegraphics[width=0.43\textwidth]{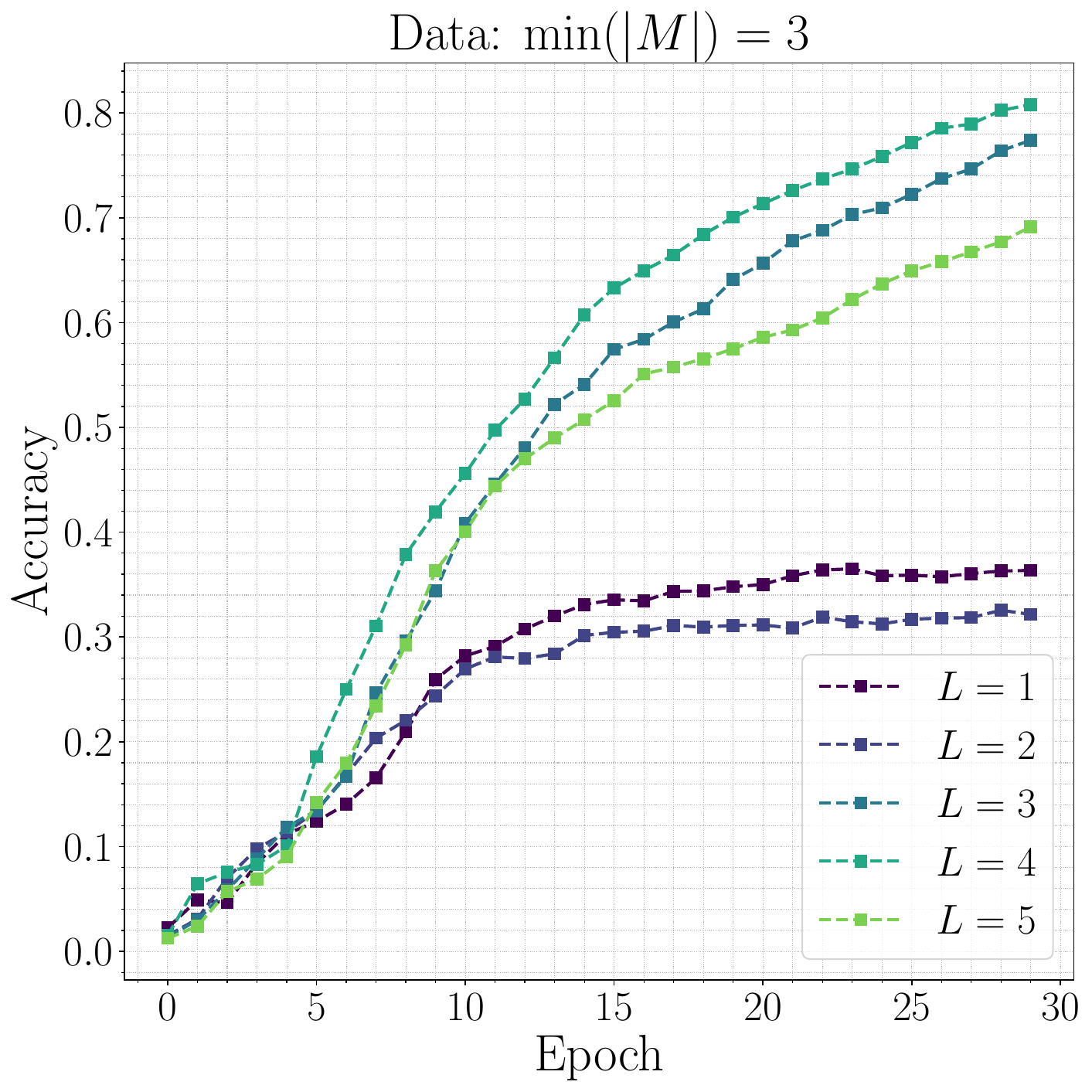}
    \vspace*{-10pt}
    \caption{Validation accuracy over epochs with different maximum message lengths ($L$) on data where $\min(|M|) = 3$. \vspace*{-10pt}}
    \label{fig:results_ms3}
\end{figure}

\paragraph{Results and Discussions.}
In order to examine the effect of the ${\min}(|M|)$ data, we run the experiments multiple times with $L = 1,2,3,4,5$ under the same setting for both data with ${\min}(|M|)=2$ and data with ${\min}(|M|)=3$ over 30 epochs. We present the results in Figure~\ref{fig:results_ms2} and \ref{fig:results_ms3}.
We see that for data with ${\min}(|M|)=2$, the difference between the accuracy of $L = 2$ and the maximum accuracy at epoch 30 is around $25\%$. However, for data with ${\min}(|M|)=3$, the difference between the accuracy of $L = 2$ and the maximum accuracy at epoch 30 is around $50\%$ which is 2 times than the data with ${\min}(|M|)=2$. This demonstrates that increasing ${\min}(|M|)=2$ for the data increases the effective message length. Additionally, increasing $L$ to be higher than ${\min}(|M|)$ is usually effective because the model tends to provide information more than minimal requirements (i.e. ${\min}(|M|)$). Moreover, in some cases, the accuracy of lower maximum $|M|$ is higher than the accuracy of higher maximum $|M|$. This is because it is easier for the model to manipulate less symbols under certain constraints. Finally, we observe that $L = 1$ can achieve to an accuracy around $35\%$ for data with ${\min}(|M|)=2$ and $L = 1, 2$ can also achieve to around $35\%$ for data with ${\min}(|M|)=3$. We hypothesize this is because we use the vocabulary size $|A| \times |V|$ and the model can learn to use 1 symbol to represent a composed phrase like "Red Triangle", by avoiding using some preset vocabularies (i.e. some of the symbols represent one word and some of the symbols represent multiple words).

\vspace{-5pt}

%% file: sections/conclusions.tex
\section{Conclusions}
In this paper, we theoretically analyzed the sampling pitfall in the training data that leads to ineffective message length of the emerged language in the Lewis signaling game. We propose the SolveMinSym \texttt{SMS} algorithm to solve the symbolic complexity for classification of the target image, and show that data with higher symbolic complexity emerge longer effective language.

%% file: sections/appendix.tex
\section{Appendix}

\subsection{Extended Related Works}
\label{appendix:ext_related_works}
Earliest research on language and communication in multi-agent setting predominantly focuses on contexts wherein agents operate with a pre-established, fixed communication language \cite{claus1998dynamics, goldman2003optimizing}. Nevertheless, this line of works do not directly address the phenomenon of language emergence, as it relies on predefined communication protocols to facilitate multi-agent cooperation.

In the pioneer research on learning the communication languages for agents, a noteworthy contribution is the model proposed by \citet{gmytrasiewicz2002toward}. This model conceptualizes negotiation as a mechanism to evolve an agent communication language from a knowledge representation language using a rule-based approach. MOCL extends classical online concept learning from single-agent to multi-agent settings for vocabulary convergence using the Perceptron algorithm \cite{wang2002mutual}. However, these works either assumes some rules pre-existing in the system or the form of language is too simple. Additionally, there's a disjoint of learning language and controlling actions.

Reinforced Inter-Agent Learning (RIAL) and Differentiable Inter-Agent Learning (DIAL) explore centralized learning coupled with decentralized execution. RIAL integrates deep Q-learning within a recurrent network framework. On the other hand, DIAL facilitates the transmission of continuous messages among agents during the centralized learning phase and transitions to discretizing these real-valued messages in the decentralized execution stage. Similarly, CommNet introduces an efficient controller designed for a variety of multi-agent reinforcement learning tasks, which enables the learning of continuous communication between agents \cite{sukhbaatar2016learning}. Both studies highlight that models equipped with learnable communication protocols demonstrate superior performance compared to those lacking such communication capabilities, but without interpretability of the emergent language. 

\subsection{\texttt{SolveMinSym (SMS)} Implementation Details}
\label{appendix:sms}

Please refer to Algorithm~\ref{alg:solve_min_sym} for the Python code.

\begin{figure*}[t!]
\noindent\hrulefill
\begin{lstlisting}[language=Python]
def SolveMinSym(target_image, all_images):
    """
    Solves the minimum number of symbols required to uniquely identify the target image
    """
    distracting_images = [img for img in all_images if img != target_image]  # Exclude target
    for combination in attribute_combinations(target_image):
        if is_unique_combination(combination, distracting_images):
            return len(combination)
    return None
    
def attribute_combinations(image):
    """
    Generates all possible non-empty combinations of the image from short to long
    """
    attrs = list(image.items())
    return chain.from_iterable(combinations(attrs, r) for r in range(1, len(attrs)+1))

def is_unique_combination(combination, distracting_images):
    """
    Tests if a given combination of attributes uniquely identifies the target image by checking if the combination of attributes in the target image can match to any distracting image
    """
    for image in distracting_images:
        match = all(image.get(key) == value for key, value in combination)
        if match:
            return False  # Found a match in distracting images (i.e. not unique)
    return True
\end{lstlisting}
\vspace{-5pt}
\noindent\hrulefill
\captionsetup{type=lstlisting}
\captionof{lstlisting}{Python code of SolveMinSym for solving ${\min}(|M|)$ \label{alg:solve_min_sym}}
\end{figure*}

\subsection{Preliminaries of Gumbel-Softmax Relaxation}
\label{appendix:gumbel}
We follow \citet{havrylov2017emergence} in using Gumbel-Softmax Relaxation for optimization. It points out that REINFORCE \cite{williams1992simple} underuses available information about the environment. Gumbel-Softmax estimator is an efficient gradient estimator that replaces the non-differentiable sample from a categorical distribution with a differentiable sample from a Gumbel-Softmax distribution \cite{jang2016categorical}.
Gumbel-Softmax Trick generates $n$-dimensional sample vectors:
\vspace*{-5pt}
\begin{equation*}
    v_i = \frac{\exp((\log \pi_i + g_i)/\tau)}{\sum_{j=1}^N \exp((\log \pi_j + g_j)/\tau)} \text{for }i = 1,...,N
\end{equation*}
where $\pi_i$ are class probabilities from a categorical distribution and $\tau$ is the temperature. However, real-value communication is not realistic compared to natural language communication. To avoid this, straight-through (GT) Gumbel-Softmax estimator discretizes $v$ using $argmax$ in the forward pass but uses continuous relaxation in the backward pass by assuming the gradient of discrete $v$ is similar to continuous $v$.